\documentclass[11pt]{article}
\usepackage[final]{acl}

\usepackage{times}
\usepackage{latexsym}
\usepackage{circledsteps}   
\usepackage{amsmath}     
\usepackage{amssymb}
\usepackage[ruled, noend]{algorithm2e}  
\usepackage{algorithmic} 
\SetAlFnt{\small}                
 
\usepackage{booktabs}       
\usepackage{graphicx}  
\usepackage{array}       
\usepackage[table]{xcolor}
\usepackage{multirow}       
\usepackage{calc}           
\usepackage[T1]{fontenc}

\usepackage{microtype}

\usepackage{inconsolata}

\usepackage{graphicx}

\title{CO-EVO: Co-evolving Semantic Anchoring and Style Diversification for Federated DG-ReID}

\author{
	Fengchun Zhang$^1$,
	Qiang Ma$^2$,
	Liuyu Xiang$^3$,
	Jinshan Lai$^1$,
	Tingxuan Huang$^4$,
		Jianwei Hu$^{2,*}$
	\\
	\vspace{-0.8em} \\
	$^1$School of Information and Software Engineering, University of Electronic Science and Technology of China \\
	$^2$QiYuan Lab \\
	$^3$School of Artificial Intelligence, Beijing University of Posts and Telecommunications \\
	$^4$School of Software, Tsinghua University
}

\begin{document}
	\maketitle
	
	{
		\renewcommand{\thefootnote}{\fnsymbol{footnote}}
		\footnotetext[1]{Corresponding author: hjw17@tsinghua.org.cn.}
	}
	
\begin{abstract}
Federated domain generalization for person re-identification (FedDG-ReID) aims to collaboratively train a pedestrian retrieval model across multiple decentralized source domains such that it can generalize to unseen target environments without compromising raw data privacy. However, this task is significantly challenged by the inherent stylistic gaps across decentralized clients. Without global supervision, models easily succumb to shortcut learning where representations overfit to domain specific camera biases rather than universal identity features. We propose CO-EVO, a novel federated framework that resolves this semantic-style conflict through a co-evolutionary mechanism. On the semantic side, Camera-Invariant Semantic Anchoring (CSA) learns identity prompts with cross-camera consistency to establish purified and domain-agnostic anchors that filter out local imaging noise. On the visual side, Global Style Diversification (GSD), powered by a Global Camera-Style Bank (GCSB), synthesizes realistic perturbations to expand the visual boundaries of training data. The core of CO-EVO is its co-evolutionary loop where purified anchors act as gravitational centers to guide the image encoder toward robust anatomical attributes amidst diverse style variations. Extensive experiments demonstrate that CO-EVO achieves state-of-the-art (SOTA) performance, proving that the synergy between semantic purification and style expansion is essential for robust cross-domain generalization. Our code is available at: \url{https://github.com/NanYiyuzurn/ACL-LGPS-2026}.
\end{abstract}

\section{Introduction}

\begin{figure}[t]
    \centering
    \includegraphics[width=0.48\textwidth]{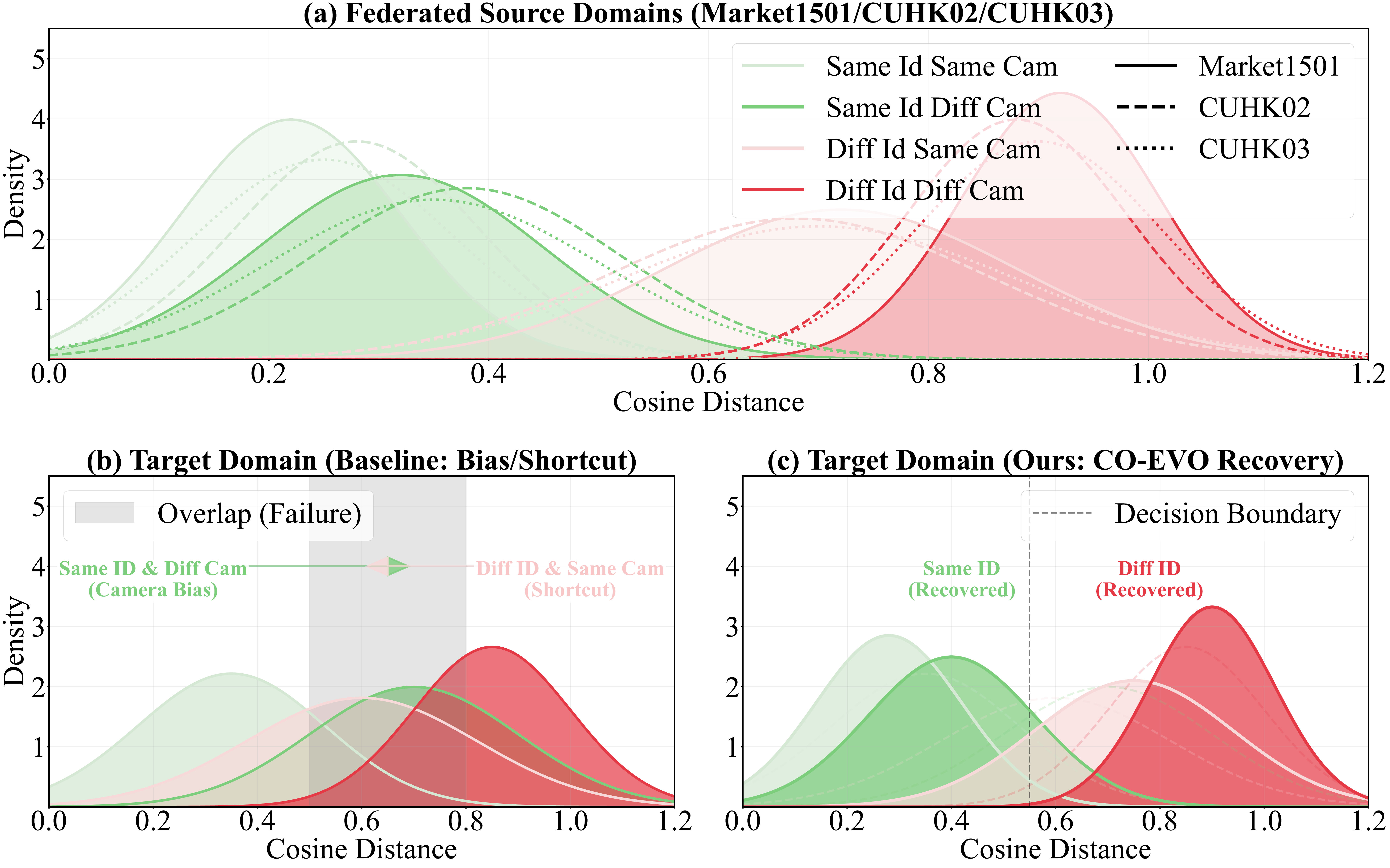}
    \caption{Cosine distance distributions illustrating the motivation of CO-EVO. (a) Source Training: Model learns identity discrimination under consistent source distributions. (b) Baseline Failure: On unseen target domains, camera bias and shortcut learning lead to distribution overlap. (c) CO-EVO Recovery: By coupling stable CSA with GSD, our framework restores the decision boundary.}
    \label{fig:motivation_dist}
\end{figure}

Person Re-identification (ReID) is a pivotal technology in modern surveillance for cross-camera pedestrian tracking and public safety \cite{luo2019bag,wang2022nformer,gao2020dcr,gao2022multigranular}. However, ReID models often face severe domain shifts in practical deployment due to heterogeneous camera characteristics and lighting conditions \cite{wang2022progressive,ye2021deep}. To address these challenges, domain generalization (DG) for ReID aims to learn a robust model from multiple sources that can generalize to unseen targets \cite{choi2021meta,dai2021generalizable,SNR,MixStyle,pAdaIN}. While traditional DG requires centralized data access, the demand for privacy has shifted focus toward Federated Learning (FL) as a paradigm for multi-domain collaboration without sharing raw data \cite{fedavg}. Consequently, Federated Domain Generalization for ReID (FedDG-ReID) has emerged to learn retrieval models from decentralized sources. Despite its potential, FedDG-ReID remains challenging due to open-set nature and client heterogeneity. Standard model aggregation often fails to achieve robust generalization as local models tend to internalize domain-specific biases.

The fundamental bottleneck of existing FedDG-ReID methods lies in a semantic-style conflict. During local optimization, the lack of a global semantic reference often leads the model toward shortcut learning. As illustrated in Figure \ref{fig:motivation_dist}(b), the network tends to exploit superficial cues such as background textures and camera related color footprints for identity discrimination. Although these cues are stable within a single client, they fail to generalize across the federation. This causes same-identity pairs from different cameras to drift apart while pulling different-identity pairs closer under similar imaging conditions. While vision-language models like CLIP show promise in anchoring visual features to stable semantic spaces, their application in FedDG-ReID is hindered by the lack of natural language names for ReID labels and high communication costs.

In parallel, local data diversification via style transfer has been adopted to mimic unseen domain shifts \cite{DACS,huang2023generalizable}. However, existing strategies face a trade-off between privacy and efficiency, often relying on costly learning-based generators that scale poorly with the number of clients \cite{FedPav,yan2020deep}. These observations raise a fundamental question: How can we achieve robust FedDG-ReID by harmonizing stable semantic grounding with efficient style diversification?

To answer this, we propose CO-EVO, a framework for Co-evolving Semantic Anchoring and Style Diversification. Here, ``co-evolving'' does not mean that semantic anchors and style templates are both updated symmetrically at every step. Instead, it denotes a coupled training mechanism in which style diversification continuously expands the visual inputs seen by the encoder, while semantic anchoring continuously constrains those updates with stable identity-level targets. As shown in Figure \ref{fig:motivation_dist}(c), CO-EVO rectifies the distribution overlap by resolving the semantic-style conflict. First, we propose Camera-Invariant Semantic Anchoring (CSA), which equips each identity with learnable tokens to form textual descriptions. Unlike previous methods, CSA introduces cross-camera consistency to distill identity-specific features from local camera noise. By caching these as frozen identity-level textual prototypes, we provide stable and purified semantic anchors that prevent the model from drifting amid visual variations. Second, we introduce Global Style Diversification (GSD) powered by a lightweight Global Camera-Style Bank (GCSB). GCSB aggregates camera statistics from all clients to generate diverse and realistic perturbations without the need for expensive generators.

The core of CO-EVO lies in this coupled optimization loop, where purified semantic anchors act as gravitational centers. These anchors guide the image encoder to focus on robust anatomical attributes even as the input visuals undergo extreme style perturbations synthesized by GSD. Our main contributions are summarized as follows:
\begin{itemize} 
    \item To the best of our knowledge, we are the first to introduce language-guided semantic supervision into the FedDG-ReID task; our framework resolves the shortcut learning problem through the proposed CSA and its distilled, camera-invariant textual anchors.
    \item We propose a GSD mechanism utilizing a global camera-style bank, enabling efficient visual diversification to simulate unseen domain shifts without violating decentralization constraints. The bank is built once with negligible overhead and remains effective even when metadata are noisy or missing.
    \item We identify and resolve the semantic-style conflict through a coupled semantic--style optimization mechanism. Extensive experiments on multiple benchmarks demonstrate that CO-EVO achieves state-of-the-art (SOTA) performance and significantly enhances cross-domain generalization. 
\end{itemize}

\section{Related Work}

\subsection{Domain Generalization for Federated Person Re-ID}
Domain Generalization (DG) for Re-ID aims to extract domain-invariant representations~\cite{Metareg,liang2025concrete}. Previous studies have explored data diversification via style perturbations to mitigate domain shifts~\cite{kang2022style,MixStyle}. In the federated learning (FL) context, existing methods such as DACS~\cite{DACS} and SSCU~\cite{SSCU} utilize a Style Transformation Model (STM) to achieve local diversification. However, these STM-based approaches require training an auxiliary generative network, which is computationally cumbersome and prone to instability during decentralized optimization. As visualized in Figure~\ref{fig:visual_evidence}(c), STM-generated images frequently suffer from repetitive artifacts and unrealistic exposure. In contrast, our CO-EVO addresses these limitations by introducing a lightweight diversification mechanism grounded in a global camera-style bank. Instead of relying on expensive generative models, we utilize template-based re-normalization of real-world style statistics to provide authentic variations with negligible overhead.

\subsection{Vision--Language Learning for Re-ID}
Vision--language models like CLIP~\cite{clip} offer powerful semantic priors through contrastive pre-training. Since Re-ID datasets lack natural language descriptions, methods like CLIP-ReID~\cite{Clip-reid} and TF-CLIP~\cite{yu2024tf} learn identity-specific prompts for semantic supervision. In federated settings, DiPrompT~\cite{bai2024diprompt} explored disentangled prompt tuning for general DG tasks. However, the synergy between vision-language semantics and federated stylization remains underexplored. Existing VLM-based methods often fail to resolve the semantic-style conflict, where aggressive stylization distorts semantic grounding. Our CO-EVO bridges this gap by establishing a coupled training loop between stable semantic anchoring and dynamic style perturbations. By decoupling the learning process, we ensure that the model consistently aligns visual features with purified, camera-invariant semantic references, even under extreme input variations.

\begin{figure*}[t]
    \centering
    \includegraphics[width=0.9\textwidth]{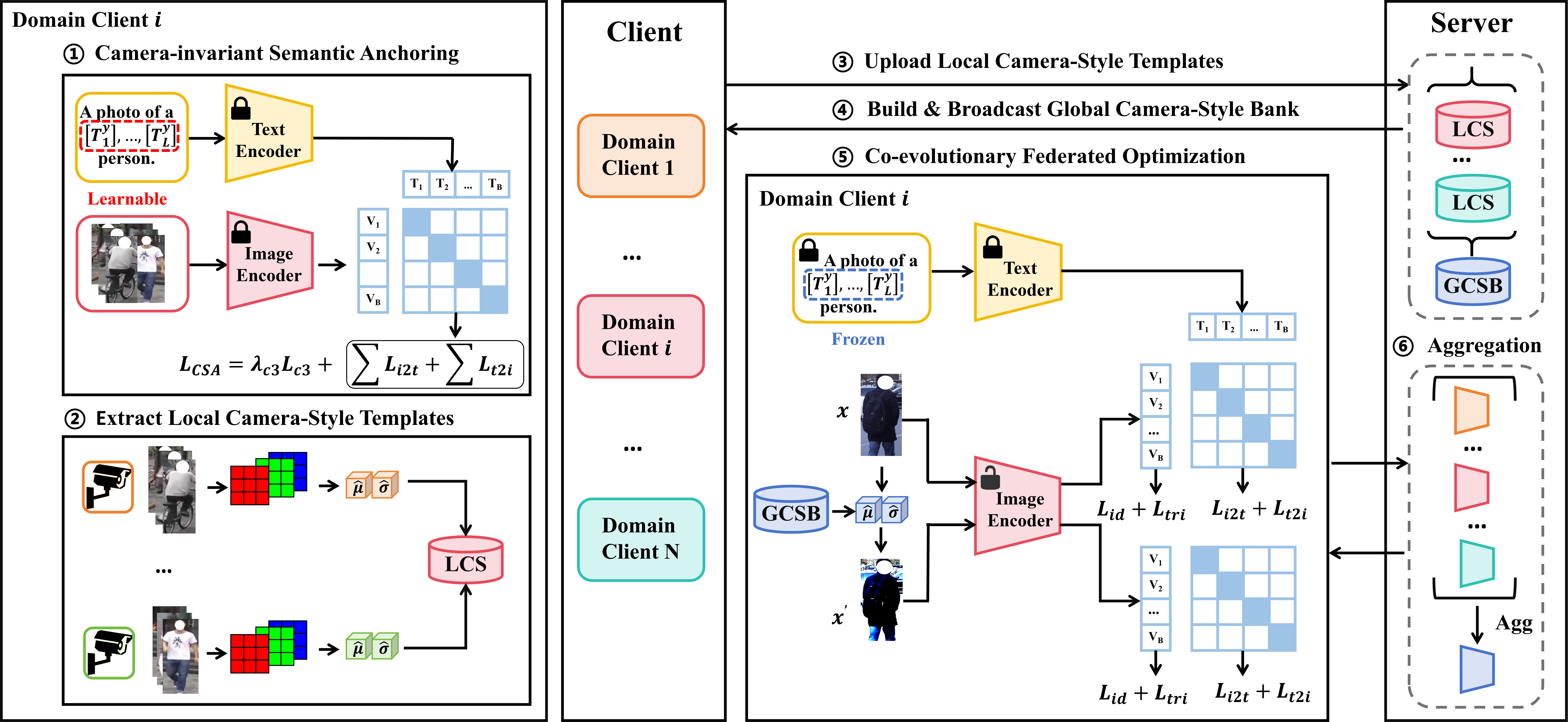}
    \caption{The overall architecture of CO-EVO for FedDG ReID. The coupled semantic--style procedure follows six key steps: \textcircled{1}--\textcircled{3} Clients establish purified semantic anchors via CSA and extract local camera-style templates; \textcircled{4} The server constructs the GCSB to broadcast; \textcircled{5} Clients perform local optimization by aligning both original and stylized views ($x, x'$) with the fixed anchors; \textcircled{6} The server aggregates local models to refine the global encoder.}
    \label{fig:pipeline}
\end{figure*}

\section{Methodology}
\label{sec:method}

\subsection{Design Rationale: The Synergy of Semantics and Style}

The core challenge in FedDG-ReID is the semantic-style conflict: purely visual supervision often succumbs to domain-specific shortcuts, while aggressive visual augmentation can corrupt identity-sensitive cues if not properly grounded. To resolve this, we propose a coupled interaction between stable semantic anchoring and dynamic style diversification. In our terminology, ``co-evolution'' specifically means that the visual distribution evolves through GSD while the encoder is repeatedly pulled back to fixed CSA anchors; it does not require simultaneous parameter updates for both branches. This mechanism compels the image encoder to map highly perturbed visual inputs back to a unified, domain-agnostic latent space.

As illustrated in Figure \ref{fig:pipeline}, we instantiate this rationale through two collaborative components: Camera-Invariant Semantic Anchoring (CSA) and Global Style Diversification (GSD). GSD forces the model to explore stylistic boundaries by synthesizing diverse camera effects grounded in global statistics, while CSA ensures that the learned representations remain anchored to intrinsic identity semantics, effectively neutralizing the impact of camera-related noise.

In our federated scenario with K clients, the non-overlapping identity sets necessitate a stable global reference to bridge domain gaps. We decouple the learning process into a stable anchoring phase and a dynamic diversification phase. By caching purified identity-level textual prototypes locally, we prevent semantic drift caused by biased local updates and eliminate the prohibitive communication cost of transmitting large-scale language models. These cached anchors serve as constant gravitational centers throughout the federated loop. As revealed in Figure \ref{fig:motivation_dist}, this coupled optimization prevents the collapse of identity distributions on unseen domains by penalizing camera-specific shortcuts while rewarding semantic-level consistency.
\subsection{Phase I: Camera-invariant Semantic Anchoring (CSA)}
\label{sec:CSA}

Most FedDG-ReID methods suffer from shortcut learning due to the lack of explicit domain-agnostic guidance. CSA resolves this by learning camera-invariant identity prompts that serve as purified semantic anchors. Unlike standard visual-language alignment, CSA explicitly distills identity-specific features from local camera noise, ensuring the resulting textual prototypes are robust across heterogeneous environments.

For each identity y on client k, we introduce L learnable tokens $\{[X_\ell^y]\}_{\ell=1}^{L}$ inserted into a template: ``a photo of a [X$_1^y$]...[X$_L^y$] person''. During this phase, we freeze the CLIP encoders, optimizing only the tokens. 

Given a mini-batch $\{(x_i,y_i,c_i)\}_{i=1}^{B}$, we adopt a bidirectional contrastive loss to align visual features $v_i$ with their corresponding textual prototypes $t_{y_i}$:
\begin{equation}
    \label{eq:i2t}
    L_{i2t}(i)= -\log\frac{\exp(s(v_i,t_{y_i})/\tau)}{\sum_{a=1}^{B}\exp(s(v_i,t_{a})/\tau)},
\end{equation}
\begin{equation}
    \small
    \label{eq:t2i}
    L_{t2i}(y)= -\frac{1}{|P(y)|}\sum_{p\in P(y)} \log\frac{\exp(s(v_p,t_{y})/\tau)}{\sum_{a=1}^{B}\exp(s(v_a,t_{y})/\tau)},
\end{equation}
where $P(y)$ denotes the set of sample indices with identity $y$, and $s(v,t)=\frac{v^\top t}{\|v\|\|t\|}$ denotes cosine similarity. To further distill the semantic information from camera-related noise, we introduce a Cross-Camera Consistency ($L_{c3}$) regularization:
\begin{equation}
    \label{eq:c3_loss}
    L_{c3} = \sum_{y \in Y_{batch}} \sum_{i,j \in P(y), c_i \neq c_j} \| s(v_i, t_y) - s(v_j, t_y) \|^2,
\end{equation}
where $c_i$ denotes the camera ID. This constraint compels the learnable tokens to ignore camera-specific visual shortcuts and focus on invariant pedestrian attributes. The overall CSA loss is $L_{CSA} = \sum L_{i2t} + \sum L_{t2i} + \lambda_{c3}L_{c3}$. After local optimization, we cache the resulting identity-level textual prototypes $T_k=\{t_y\}_{y\in Y_k}$ as purified semantic anchors for the federated loop:
\begin{equation}
    \label{eq:text-proto}
    T_k=\{t_y\}_{y\in Y_k}\in R^{|Y_k|\times D}.
\end{equation}

\subsection{Phase II: Global Style Diversification (GSD)}
\label{sec:gsd}

While CSA provides purified and camera-invariant semantic anchors, the image encoder must still encounter a vast spectrum of visual variations to achieve robust domain invariance. To this end, we propose Global Style Diversification (GSD), a lightweight mechanism to synthesize realistic cross-client domain shifts.

Central to GSD is the Global Camera-Style Bank (GCSB), which serves as a repository of camera-specific style statistics. We use channel-wise mean and variance because prior DG studies have shown that these first- and second-order feature statistics capture domain-specific appearance factors such as illumination, color tone, and texture while largely preserving semantic structure~\cite{MixStyle,CrossStyle}. For each client $k$ and camera $c \in C_k$, we extract a camera-style template by computing channel-wise statistics:
\begin{equation}
    \label{eq:style-template}
    (\mu_{k,c},\sigma^2_{k,c})=Stat(\{x_i^k~|~c_i^k=c\}).
\end{equation}
The server aggregates these templates into a global repository $\mathcal{B}=\cup_{k=1}^{K}\cup_{c\in C_k}\{(\mu_{k,c},\sigma^2_{k,c})\}$. When camera IDs are unreliable or unavailable, we can replace $c$ with pseudo-groups obtained from unsupervised clustering, allowing GSD to remain applicable without changing the training objective. During local optimization, we inject diverse camera effects via template-based re-normalization:
\begin{equation}
    \label{eq:stylize}
    \hat{x}=\frac{x-\mu(x)}{\sqrt{\sigma^2(x)+\epsilon}},\qquad
    x'=\hat{x}\odot \sqrt{\sigma_s^2}+\mu_s,
\end{equation}
where $(\mu_s,\sigma^2_s) \sim \mathcal{B}$. This ensures that the augmented view $x'$ is grounded in real camera distributions rather than arbitrary noise. In practice, the GCSB is constructed only once before federated optimization, takes about 4s per client on average, introduces no additional trainable parameters, and accounts for less than $0.1\%$ of the full training time. We emphasize that GSD mainly targets photometric variation rather than explicit geometric transformation. Robustness to viewpoint and scale changes instead comes from the CLIP backbone's transferable priors and the cross-camera semantic constraint imposed by CSA.

\begin{algorithm}[t]
    \LinesNumbered
    \caption{CO-EVO: Coupled Federated Learning with Stable Anchors and Style Diversification}
    \label{alg:coevo}
    \KwIn{Clients $\{D_k\}_{k=1}^{K}$; rounds R; local epochs E; weight $\lambda_{c3}$; weight $\lambda$.}
    \KwOut{Global image encoder $\theta^{R}$.}
    
    \textbf{Phase I: Camera-Invariant Semantic Anchoring (CSA)}\\
    \For{$k=1,\dots,K$ in parallel}{
        Optimize local identity tokens via $L_{CSA}$ (Eq.~\ref{eq:i2t}-\ref{eq:c3_loss})\;
        Cache purified prototypes $T_k = \{t_y\}_{y\in Y_k}$ (Eq.~\ref{eq:text-proto})\;
        Upload style templates $\{(\mu, \sigma^2)\}$ (Eq.~\ref{eq:style-template}) to server\;
    }
    
    \textbf{Phase II \& III: Coupled Training Loop}\\
    Server constructs GCSB $\mathcal{B}$ and broadcasts $\mathcal{B}, \theta^0$ to all clients\;
    \For{$r=1,\dots,R$}{
        \For{$k=1,\dots,K$ in parallel}{
            Set local model $\theta_k \leftarrow \theta^{r-1}$\;
            \For{$e=1,\dots,E$}{
                Sample $(x,y) \sim D_k$ and synthesize $x' \sim GSD(\mathcal{B})$ (Eq.~\ref{eq:stylize})\;
                Compute $L_{loc}$ (Eq.~\ref{eq:local-obj}) for original and stylized views\;
                Update $\theta_k$ via backpropagation\;
            }
            Upload $\theta_k^{r}$ to server\;
        }
        $\theta^{r} \leftarrow \sum_{k=1}^{K} \frac{n_k}{N} \theta_k^{r}$ and broadcast $\theta^{r}$ to all clients\;
    }
    \Return{$\theta^{R}$}\;
\end{algorithm}

\subsection{Coupled Federated Optimization}
\label{sec:fedopt}

The core of CO-EVO lies in the joint optimization of semantic stability (via CSA) and visual diversity (via GSD) within the federated loop. During local training on client $k$, we sample a mini-batch $\{(x_i,y_i)\}$ and synthesize stylized counterparts $x'_i$ using templates from the GCSB. This is the operational meaning of our ``co-evolution'' terminology: the input distribution evolves through sampled style templates, while the encoder parameters evolve under constant semantic anchors. To ensure discriminability, we apply identity loss $L_{id}$ and triplet loss $L_{tri}$ to both original ($x$) and stylized ($x'$) views:
\begin{equation}
    \label{eq:lid}
    L_{id}(\tilde{x}) = -\frac{1}{B}\sum_{i=1}^{B} \log p_{i,y_i}(\tilde{x}),
\end{equation}
\begin{equation}
    \label{eq:ltri}
    L_{tri}(\tilde{x}) = \frac{1}{B}\sum_{i=1}^{B}\max(d_p^{(i)}(\tilde{x}) - d_n^{(i)}(\tilde{x}) + \alpha, 0),
\end{equation}
where $\tilde{x} \in \{x, x'\}$. To prevent the model from exploiting domain-specific shortcuts, we utilize the purified textual prototypes $T_k$ as fixed anchors. For each view $\tilde{x}$, the semantic alignment loss is defined as:
\begin{equation}
    \label{eq:li2tce}
    L_{align}(i;\tilde{x}) = -\log \frac{\exp(s(v_i(\tilde{x}),t_{y_i})/\tau)}{\sum_{y\in Y_k}\exp(s(v_i(\tilde{x}),t_{y})/\tau)}.
\end{equation}
By forcing both x and x' to align with the same camera-invariant anchor $t_{y_i}$, the image encoder is compelled to discard low-level stylistic noise. The total local objective is:
\begin{equation}
    \label{eq:local-obj}
    L_{loc} = \sum_{\tilde{x} \in \{x, x'\}} ( L_{id}(\tilde{x}) + L_{tri}(\tilde{x}) + \lambda L_{align}(\tilde{x}) ).
\end{equation}
This coupled process ensures that while GSD expands visual boundaries, the CSA anchors provide a consistent gravitational center that restores the decision boundary, as visualized in Figure~\ref{fig:motivation_dist}. After $E$ local epochs, the server performs weighted aggregation to update the global model $\theta^{r} = \sum_{k=1}^{K} \frac{n_k}{N} \theta_k^{r}$ and synchronizes the GCSB to incorporate evolving camera statistics. The complete procedure is summarized in Algorithm~\ref{alg:coevo}.

\begin{table*}[t]
	\centering
	\caption{Protocol~I (leave-one-domain-out) results on FedDG-ReID. Each source domain is treated as a client. We report mAP and Rank-1 (\%) on three held-out target domains and the average.}
	\resizebox{0.9\textwidth}{!}{
		\begin{tabular}{l l l c c c c c c c c}
			\toprule
			Category & Methods & Reference & \multicolumn{2}{c}{MS+C2+C3$\rightarrow$M} & \multicolumn{2}{c}{M+C2+C3$\rightarrow$MS} & \multicolumn{2}{c}{MS+C2+M$\rightarrow$C3} & \multicolumn{2}{c}{Average} \\
			\cmidrule(lr){4-5} \cmidrule(lr){6-7} \cmidrule(lr){8-9} \cmidrule(lr){10-11}
			& & & mAP & rank-1 & mAP & rank-1 & mAP & rank-1 & mAP & rank-1 \\
			\midrule
			Federated Learning & SCAFFOLD & ICML 2020 & 26.0 & 50.5 & 5.3 & 15.8 & 22.9 & 26.0 & 18.1 & 30.8 \\
			& MOON & CVPR 2021 & 26.8 & 51.1 & 4.8 & 14.5 & 20.9 & 22.5 & 17.5 & 29.4 \\
			& FedProx & MLSYS 2021 & 29.3 & 53.8 & 5.8 & 17.4 & 19.1 & 17.7 & 18.1 & 29.7 \\
			\midrule
			Domain Generalization & MixStyle & ICLR 2020 & 31.2 & 53.5 & 5.5 & 16.0 & 28.6 & 31.5 & 21.8 & 33.6 \\
			& CrossStyle & ICCV 2021 & 35.5 & 59.6 & 4.6 & 14.0 & 27.8 & 28.0 & 22.6 & 33.9 \\
			\midrule
			Federated-ReID & FedReID & AAAI 2021 & 30.1 & 53.7 & 4.5 & 13.7 & 26.4 & 26.5 & 20.3 & 31.3 \\
			& FedPav & MM 2020 & 25.4 & 49.4 & 5.2 & 15.5 & 22.5 & 24.3 & 17.7 & 29.7 \\
			\midrule
			DG-ReID & SNR & CVPR 2020 & 32.7 & 59.4 & 5.1 & 15.3 & 28.5 & 30.0 & 22.1 & 34.9 \\
			\midrule
			FedDG-ReID (RN50) & DACS & AAAI 2024 & 36.3 & 61.2 & 10.4 & 27.5 & 30.7 & 34.1 & 25.8 & 40.9 \\
			& SSCU & MM 2025 & 39.5 & 66.4 & 11.9 & 32.3 & 32.8 & 34.1 & 28.1 & 44.3 \\
			& \cellcolor{gray!20}CO-EVO & \cellcolor{gray!20}ours & \cellcolor{gray!20}\textbf{42.4} & \cellcolor{gray!20}\textbf{71.2} & \cellcolor{gray!20}\textbf{12.9} & \cellcolor{gray!20}\textbf{33.7} & \cellcolor{gray!20}\textbf{34.9} & \cellcolor{gray!20}\textbf{37.1} & \cellcolor{gray!20}\textbf{30.1} & \cellcolor{gray!20}\textbf{47.3} \\
			\midrule
			FedDG-ReID (ViT) & FedPav (ViT) & MM 2020 & 37.4 & 62.6 & 14.6 & 33.7 & 23.7 & 25.0 & 25.2 & 40.4 \\
			& CrossStyle (ViT) & ICCV 2021 & 41.4 & 65.8 & 17.9 & 40.8 & 31.0 & 38.4 & 30.1 & 48.3 \\
			& DACS (ViT) & AAAI 2024 & 45.4 & 70.7 & 20.3 & 44.2 & 36.6 & 42.1 & 34.1 & 52.3 \\
			& \cellcolor{gray!20}CO-EVO & \cellcolor{gray!20}ours & \cellcolor{gray!20}\textbf{60.7} & \cellcolor{gray!20}\textbf{80.2} & \cellcolor{gray!20}\textbf{32.2} & \cellcolor{gray!20}\textbf{60.3} & \cellcolor{gray!20}\textbf{51.3} & \cellcolor{gray!20}\textbf{52.7} & \cellcolor{gray!20}\textbf{48.1} & \cellcolor{gray!20}\textbf{64.4} \\
			\bottomrule
		\end{tabular}
	}
	\label{tab:reid_performance}
\end{table*}

\section{Experiments}

\subsection{Experimental Settings}
In this section, we present our experimental setup, focusing on the datasets and representative baselines used for evaluation. Other detailed configurations, including the federated setup, backbone architectures, evaluation protocols, and specific implementation details, are provided in the Appendix.

\paragraph{Datasets.}
We conduct experiments on four large-scale person ReID benchmarks: CUHK02 \cite{cuhk02}, CUHK03 \cite{cuhk03}, MSMT17 \cite{msmt}, and Market1501 \cite{market}. For clarity, these domains are denoted as C2, C3, MS, and M, respectively.

\paragraph{Baselines.}
We evaluate CO-EVO against representative methods from four categories:
(i) Generic federated optimization algorithms including SCAFFOLD \cite{Scaffold}, MOON \cite{moon}, and FedProx \cite{fedprox}, which address client drift and heterogeneity. 
(ii) Style-based domain generalization techniques such as MixStyle \cite{MixStyle} and CrossStyle \cite{CrossStyle} that diversify distributions via style statistics.
(iii) Federated ReID frameworks like FedPav \cite{FedPav} and FedReID \cite{FedReID}, which are tailored for open-set retrieval.
(iv) Specialized DG-ReID and FedDG-ReID methods, including SNR \cite{SNR}, DACS \cite{DACS}, and SSCU \cite{SSCU}. Among them, DACS and SSCU are the current SOTA for FedDG-ReID.
For a fair comparison, all baselines are trained under the same federated setup (domain-as-client), backbone, and communication rounds.

\begin{table}[t]
	\centering
	\caption{Protocol~II results with a reduced number of source domains while keeping MS as a source client. We report mAP and Rank-1 (\%) on targets M and C3 under different source combinations.}
	\resizebox{0.5\textwidth}{!}{
		\begin{tabular}{l c c c c c c}
			\toprule
			Methods & \multicolumn{2}{c}{MS+C3$\rightarrow$M} & \multicolumn{2}{c}{MS+C2$\rightarrow$M} & \multicolumn{2}{c}{MS+C2+C3$\rightarrow$M} \\
			\cmidrule(lr){2-3} \cmidrule(lr){4-5} \cmidrule(lr){6-7}
			& mAP & rank-1 & mAP & rank-1 & mAP & rank-1 \\
			\midrule
			FedPav & 27.5 & 51.5 & 24.8 & 48.5 & 25.4 & 49.1 \\
			FedReID & 31.0 & 55.0 & 28.1 & 52.4 & 30.1 & 53.7 \\
			DACS & 33.2 & 58.1 & 30.3 & 56.3 & 36.3 & 61.2 \\
			SSCU & 36.7 & 62.8 & 34.8 & 62.7 & 39.5 & 66.4 \\
			\cellcolor{gray!20}ours & \cellcolor{gray!20}\textbf{39.3} & \cellcolor{gray!20}\textbf{68.4} & \cellcolor{gray!20}\textbf{36.9} & \cellcolor{gray!20}\textbf{67.1} & \cellcolor{gray!20}\textbf{42.4} & \cellcolor{gray!20}\textbf{71.2} \\
			\midrule\midrule
			Methods & \multicolumn{2}{c}{MS+M$\rightarrow$C3} & \multicolumn{2}{c}{MS+C2$\rightarrow$C3} & \multicolumn{2}{c}{MS+C2+M$\rightarrow$C3} \\
			\cmidrule(lr){2-3} \cmidrule(lr){4-5} \cmidrule(lr){6-7}
			& mAP & rank-1 & mAP & rank-1 & mAP & rank-1 \\
			\midrule
			FedPav & 15.2 & 14.1 & 17.3 & 17.0 & 22.5 & 24.3 \\
			FedReID & 16.1 & 15.3 & 21.8 & 20.4 & 26.4 & 26.5 \\
			DACS & 18.2 & 17.7 & 22.9 & 23.5 & 30.7 & 34.1 \\
			SSCU & 20.9 & 20.8 & 27.1 & 29.3 & 32.8 & 34.1 \\
			\cellcolor{gray!20}ours & \cellcolor{gray!20}\textbf{24.3} & \cellcolor{gray!20}\textbf{25.1} & \cellcolor{gray!20}\textbf{29.4} & \cellcolor{gray!20}\textbf{34.2} & \cellcolor{gray!20}\textbf{34.9} & \cellcolor{gray!20}\textbf{37.1} \\
			\bottomrule
		\end{tabular}
	}
	\label{tab:transfer_performance}
\end{table}

\begin{table}[t]
	\centering
	\caption{Protocol~III (source-domain evaluation) results. Models are trained with clients \{M, C2, C3\} and evaluated on the test split of each source domain. We report mAP and Rank-1 (\%).}
	\resizebox{0.5\textwidth}{!}{
		\begin{tabular}{l c c c c c c}
			\toprule
			Methods & \multicolumn{2}{c}{M+C2+C3$\rightarrow$M} & \multicolumn{2}{c}{M+C2+C3$\rightarrow$C2} & \multicolumn{2}{c}{M+C2+C3$\rightarrow$C3} \\
			\cmidrule(lr){2-3} \cmidrule(lr){4-5} \cmidrule(lr){6-7}
			& mAP & rank-1 & mAP & rank-1 & mAP & rank-1 \\
			\midrule
			FedProx & 61.0 & 80.4 & 66.8 & 65.5 & 24.2 & 23.9 \\
			FedPav & 53.9 & 76.0 & 59.7 & 56.3 & 19.6 & 19.6 \\
			FedReID & 71.8 & 87.6 & 82.9 & 82.8 & 44.0 & 44.9 \\
			DACS & 72.1 & 88.2 & 84.5 & 83.4 & 47.4 & 50.1 \\
			SSCU & 73.0 & 88.7 & 84.9 & 83.9 & 50.4 & 53.2 \\
			\cellcolor{gray!20}ours & \cellcolor{gray!20}\textbf{81.4} & \cellcolor{gray!20}\textbf{93.1} & \cellcolor{gray!20}\textbf{89.8} & \cellcolor{gray!20}\textbf{91.2} & \cellcolor{gray!20}\textbf{59.7} & \cellcolor{gray!20}\textbf{63.1} \\
			\bottomrule
		\end{tabular}
	}
	\label{tab:m_c2_c3_transfer}
\end{table}

\subsection{Comparison under Three Evaluation Protocols} 
\label{sec:exp_protocols}

\paragraph{Protocol I: Leave-One-Domain-Out.} 
Table~\ref{tab:reid_performance} summarizes the generalization results under the most rigorous DG setting. CO-EVO achieves SOTA performance across all benchmarks, outperforming the strongest CNN-based baseline by an average of +2.0\% mAP and +3.0\% Rank-1 (28.1/44.3$\rightarrow$30.1/47.3). Notably, on the most challenging MSMT17 domain, our method obtains the best results (12.9/33.7). We attribute this to the coupled effect of CSA and GSD: while CSA distills purified semantic anchors via cross-camera consistency, GSD forces the model to maintain these anchors amid extreme stylistic shifts. Consistent gains under Transformer backbones further verify the architectural robustness of our design.

\paragraph{Protocol II: Scaling with Source Domains.} 
Table~\ref{tab:transfer_performance} evaluates generalization under varying numbers of source clients. While performance generally improves as more domains participate, CO-EVO consistently maintains its lead under all source combinations. For instance, on Market-1501, the performance scales from 39.3\% mAP (MS+C3) to 42.4\% mAP (MS+C2+C3). This suggests that even with limited source diversity, our camera-invariant semantic anchors provide essential structural constraints that prevent the model from overfitting to specific local camera styles, a common pitfall in standard federated ReID.

\paragraph{Protocol III: Source-Domain Discriminability.} 
Table~\ref{tab:m_c2_c3_transfer} reports the testing results on participating source domains. CO-EVO achieves the best performance across all clients, indicating that the pursuit of domain generalization does not compromise the model's discriminative power on local data. By resolving the semantic-style conflict, the learned representations successfully internalize robust identity cues that are both domain-agnostic for unseen targets and highly discriminative for participating clients. This dual-advantage confirms that our purified semantic grounding effectively captures intrinsic biometric features rather than superficial imaging shortcuts.

\subsection{Robustness to Imperfect or Missing Metadata}
\label{sec:metadata_robustness}

\begin{table}[t]
    \centering
    \small
    \caption{Robustness of CO-EVO under imperfect camera metadata. The clean SSCU baseline is included for reference.}
    \label{tab:metadata_robustness}
    \resizebox{1.0\linewidth}{!}{
    \begin{tabular}{l | l | cc | cc | cc}
        \toprule
        Metadata & Method / Scope & \multicolumn{2}{c|}{MS+C2+C3$\rightarrow$M} & \multicolumn{2}{c|}{M+C2+C3$\rightarrow$MS} & \multicolumn{2}{c}{MS+C2+M$\rightarrow$C3} \\
        Setting &  & mAP & R1 & mAP & R1 & mAP & R1 \\
        \midrule
         \rowcolor{gray!15}
        Clean & CO-EVO & \textbf{42.4} & \textbf{71.2} & \textbf{12.9} & \textbf{33.7} & \textbf{37.1} & \textbf{38.9} \\
        Noisy & 30\% Camera ID Noise & 40.4 & 68.7 & 11.7 & 32.4 & 34.6 & 36.2 \\
        Missing & K-means Pseudo-Groups & 41.1 & 69.8 & 12.2 & 33.1 & 35.8 & 37.5 \\
        Reference & SSCU (clean labels) & 39.5 & 66.4 & 11.9 & 32.3 & 32.8 & 34.1 \\
        \bottomrule
    \end{tabular}}
\end{table}

\paragraph{Noisy, Missing, and Unreliable Metadata.}
Table~\ref{tab:metadata_robustness} shows that CO-EVO remains effective even when the metadata used to build the GCSB are imperfect. Under 30\% camera-ID corruption, the performance drops are moderate, indicating that the shared style bank is not overly brittle to annotation noise. In the zero-metadata case, we replace camera IDs with K-means pseudo-groups and still outperform SSCU trained with clean labels on all three transfers. These results suggest that GSD benefits from coarse grouping structure rather than perfect camera annotations: as long as the pseudo-groups preserve dominant appearance patterns, the resulting templates remain useful for expanding the visual support seen during training. This robustness is particularly important in federated deployments, where camera identifiers may be incomplete, noisy, or inconsistent across institutions.

\subsection{Hyperparameter Analysis}
\label{sec:hyperparameter}

\begin{table}[h]
    \centering
    \small
    \caption{Impact of token length $L$ in CSA. Results follow Protocol I (Leave-one-domain-out).}
    \label{tab:csa_tokens}
    \resizebox{1.0\linewidth}{!}{
    \begin{tabular}{c | cc | cc | cc}
        \toprule
        $L$ & \multicolumn{2}{c|}{MS+C2+C3$\rightarrow$M} & \multicolumn{2}{c|}{M+C2+C3$\rightarrow$MS} & \multicolumn{2}{c}{MS+C2+M$\rightarrow$C3} \\
        & mAP & R1 & mAP & R1 & mAP & R1 \\
        \midrule
        1  & 40.8 & 69.4 & 10.3 & 29.5 & 35.5 & 36.8 \\
        \rowcolor{gray!15}
        4  & \textbf{42.4} & \textbf{71.2} & \textbf{12.9} & \textbf{33.7} & \textbf{37.1} & \textbf{38.9} \\
        8  & 41.9 & 70.7 & 12.1 & 32.8 & 36.6 & 38.2 \\
        16 & 41.5 & 70.2 & 11.5 & 31.6 & 36.2 & 37.7 \\
        \bottomrule
    \end{tabular}}
\end{table}

\paragraph{Impact of Token Length ($L$).}
We evaluate the sensitivity of token length $L \in \{1, 4, 8, 16\}$ in Table~\ref{tab:csa_tokens}. Performance consistently peaks at $L=4$ across all transfer tasks. While $L=1$ is too generic to capture identity nuances, larger values ($L \geq 8$) introduce redundant parameters that overfit to local camera-specific noise, leading to semantic drift. A compact set of 4 tokens provides the optimal balance between discriminative power and anchoring stability for the coupled semantic--style optimization.

\paragraph{Sensitivity of Cross-Camera Consistency ($\lambda_{c3}$).}
We investigate the impact of the consistency weight $\lambda_{c3}$ in Table~\ref{tab:c3_sensitivity}. When $\lambda_{c3}=0$, the textual prototypes are prone to capturing local camera footprints, leading to sub-optimal generalization on unseen domains. As $\lambda_{c3}$ increases to 0.1, the performance improves significantly, particularly on the complex MSMT17 domain (+1.4\% mAP). This gain validates that $\mathcal{L}_{c3}$ effectively ``purifies'' the semantic anchors by filtering out camera-specific noise during the anchoring phase. When $\lambda_{c3}$ becomes too large, however, the constraint starts to over-suppress subtle cross-camera appearance differences that are still useful for fine-grained identity discrimination, which explains the mild degradation beyond 0.1.

\begin{table}[h]
    \centering
    \small
    \caption{Sensitivity analysis of the cross-camera consistency weight $\lambda_{c3}$ in CSA.}
    \label{tab:c3_sensitivity}
    \resizebox{1.0\linewidth}{!}{
    \begin{tabular}{c | cc | cc | cc}
        \toprule
        $\lambda_{c3}$ & \multicolumn{2}{c|}{MS+C2+C3$\rightarrow$M} & \multicolumn{2}{c|}{M+C2+C3$\rightarrow$MS} & \multicolumn{2}{c}{MS+C2+M$\rightarrow$C3} \\
        & mAP & R1 & mAP & R1 & mAP & R1 \\
        \midrule
        0.0  & 41.2 & 69.8 & 11.5 & 31.8 & 35.8 & 37.5 \\
        \rowcolor{gray!15}
        0.1  & \textbf{42.4} & \textbf{71.2} & \textbf{12.9} & \textbf{33.7} & \textbf{37.1} & \textbf{38.9} \\
        0.2  & 42.1 & 70.9 & 12.4 & 33.1 & 36.7 & 38.3 \\
        0.5  & 41.5 & 70.2 & 11.8 & 32.0 & 36.2 & 37.8 \\
        \bottomrule
    \end{tabular}}
\end{table}

\subsection{Ablation Study}
\label{sec:ablation}

We conduct systematic ablations under the CO-EVO framework to quantify the individual and synergistic gains of its core modules.

\begin{table}[h]
    \centering
    \small
    \caption{Ablation of key components: CSA (Phase I) and GSD (Phase II).}
    \label{tab:ablation_main}
    \resizebox{1.0\linewidth}{!}{
    \begin{tabular}{cc | cc | cc | cc}
        \toprule
        \multicolumn{2}{c|}{Components} & \multicolumn{2}{c|}{MS+C2+C3$\rightarrow$M} & \multicolumn{2}{c|}{M+C2+C3$\rightarrow$MS} & \multicolumn{2}{c}{MS+C2+M$\rightarrow$C3} \\
        CSA & GSD & mAP & R1 & mAP & R1 & mAP & R1 \\
        \midrule
        - & - & 25.4 & 49.4 & 5.2 & 15.5 & 22.5 & 24.3 \\
        \checkmark & - & 38.7 & 66.7 & 9.1 & 30.7 & 33.8 & 35.2 \\
        - & \checkmark & 39.8 & 67.6 & 10.8 & 30.3 & 34.1 & 35.9 \\
        \rowcolor{gray!15}
        \checkmark & \checkmark & \textbf{42.4} & \textbf{71.2} & \textbf{12.9} & \textbf{33.7} & \textbf{37.1} & \textbf{38.9} \\
        \bottomrule
    \end{tabular}}
\end{table}

\paragraph{Synergy of Semantics and Style.} 
Table~\ref{tab:ablation_main} demonstrates that the vanilla federated baseline fails significantly under severe domain shifts (5.2\% mAP on MSMT17). 
(1) CSA Stability: CSA provides stable, purified semantic anchors, preventing identity features from drifting across heterogeneous clients. 
(2) GSD Diversity: GSD mitigates camera bias by diversifying training distributions via global statistics, breaking visual shortcuts. 
(3) Full Synergy: CO-EVO achieves optimal performance, confirming that CSA ensures semantic grounding while GSD explores diverse style boundaries.

\begin{table}[h]
    \centering
    \small
    \caption{Ablation of CSA anchoring strategy. Stat.: Static Caching (Ours).}
    \label{tab:csa_fine}
    \resizebox{1.0\linewidth}{!}{
    \begin{tabular}{l | cc | cc | cc}
        \toprule
        Strategy & \multicolumn{2}{c|}{MS+C2+C3$\rightarrow$M} & \multicolumn{2}{c|}{M+C2+C3$\rightarrow$MS} & \multicolumn{2}{c}{MS+C2+M$\rightarrow$C3} \\
        & mAP & R1 & mAP & R1 & mAP & R1 \\
        \midrule
        Baseline & 25.4 & 49.4 & 5.2 & 15.5 & 22.5 & 24.3 \\
        CSA (Dyn.) & 39.1 & 67.8 & 10.4 & 30.1 & 34.2 & 35.6 \\
        \rowcolor{gray!15}
        CSA (Stat.) & \textbf{42.4} & \textbf{71.2} & \textbf{12.9} & \textbf{33.7} & \textbf{37.1} & \textbf{38.9} \\
        \bottomrule
    \end{tabular}}
\end{table}

\paragraph{Semantic Anchoring Strategy.}
Table~\ref{tab:csa_fine} investigates prototype stability. Compared to dynamic updates, our \textit{Static Caching} achieved higher accuracy, confirming that dynamic prompts are prone to inheriting local camera noise, whereas frozen, purified anchors provide a reliable ``gravitational center'' for cross-modal alignment.

\begin{table}[h]
    \centering
    \small
    \caption{Ablation of GSD sampling scope. Glob.: Global-Bank (Ours).}
    \label{tab:gsd_fine}
    \resizebox{1.0\linewidth}{!}{
    \begin{tabular}{l | cc | cc | cc}
        \toprule
        Scope & \multicolumn{2}{c|}{MS+C2+C3$\rightarrow$M} & \multicolumn{2}{c|}{M+C2+C3$\rightarrow$MS} & \multicolumn{2}{c}{MS+C2+M$\rightarrow$C3} \\
        & mAP & R1 & mAP & R1 & mAP & R1 \\
        \midrule
        Baseline & 38.7 & 66.7 & 9.1 & 30.7 & 33.8 & 35.2 \\
        Random-Stat & 39.1 & 67.2 & 9.7 & 30.9 & 34.5 & 35.6 \\
        GSD (Loc.) & 40.2 & 68.5 & 10.5 & 31.4 & 35.1 & 36.3 \\
        \rowcolor{gray!15}
        GSD (Glob.) & \textbf{42.4} & \textbf{71.2} & \textbf{12.9} & \textbf{33.7} & \textbf{37.1} & \textbf{38.9} \\
        \bottomrule
    \end{tabular}}
\end{table}

\paragraph{GSD Sampling Scope.}
Table~\ref{tab:gsd_fine} compares stylization scopes. Randomly perturbing feature statistics yields only marginal gains over the baseline, which suggests that naive noise injection is insufficient to approximate realistic deployment shifts. In contrast, the Global-Bank strategy yields significantly better results, validating that sharing low-dimensional camera statistics effectively constructs a proxy distribution of unseen domains and that the improvement of GSD comes from real camera-driven templates rather than arbitrary perturbations.

\begin{figure*}[t]
    \centering
    \includegraphics[width=0.8\linewidth]{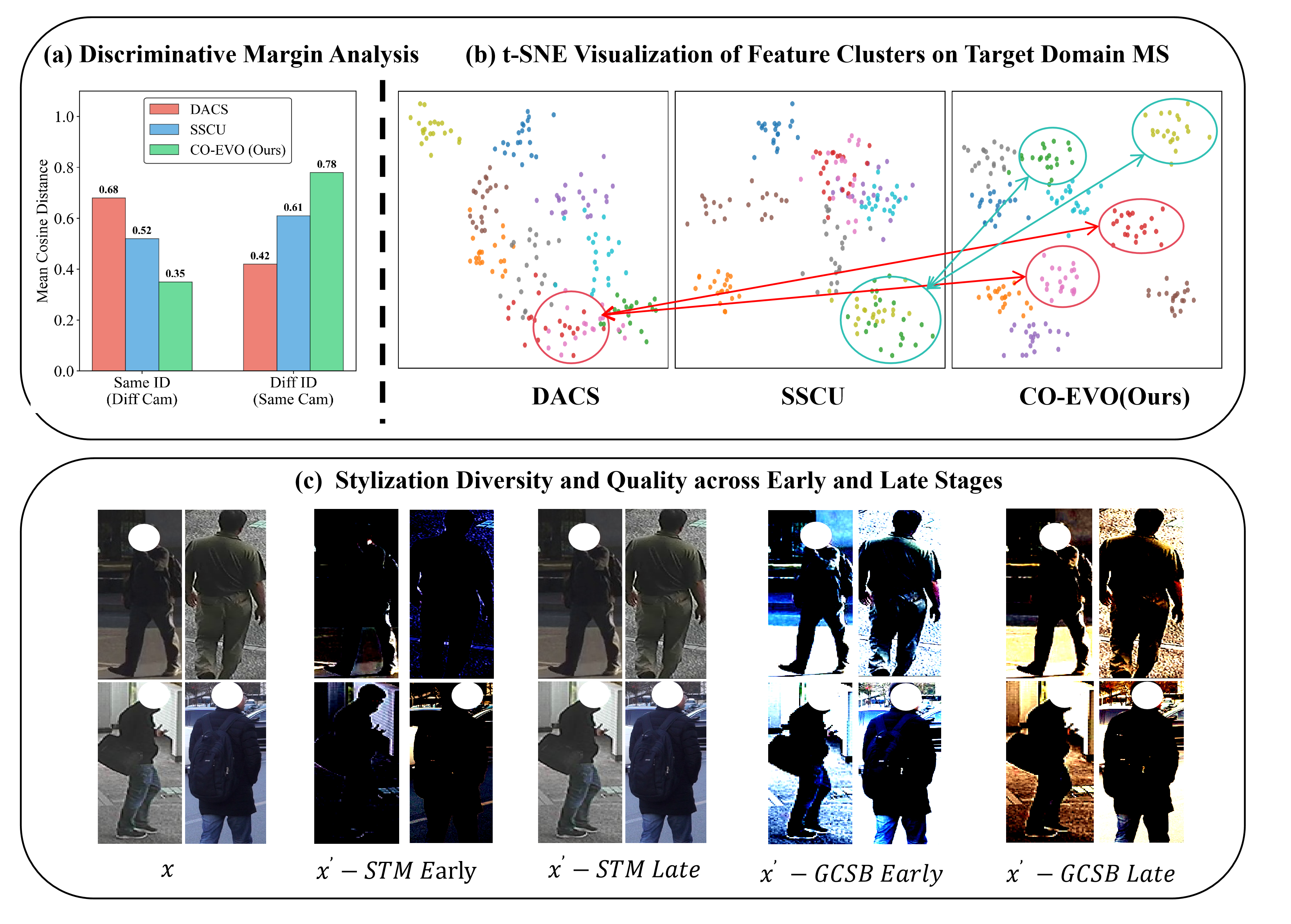}
    \caption{Comprehensive evaluation of the proposed CO-EVO. 
    (a) Discriminative Margin Analysis: Quantitative comparison of mean cosine distances for same-identity and different-identity pairs. 
    (b) t-SNE Performance on Target Domain MS: Visualization of feature clusters on the unseen MS domain (models trained on M+C2+C3), showing the evolution from DACS and SSCU to our CO-EVO. 
    (c) Stylization Diversity and Quality across Early and Late Stages: Qualitative comparison between the baseline STM and our GCSB-based GSD module during the training process.}
    \label{fig:visual_evidence}
\end{figure*}

\subsection{Visual and Statistical Insights}
\label{sec:visual_analysis}

To delve deeper into the mechanism of CO-EVO, we provide comprehensive visualizations and statistical evidence in Figure~\ref{fig:visual_evidence}. Quantitatively, Figure~\ref{fig:visual_evidence}(a) highlights the restoration of the discriminative margin. By leveraging CSA anchors, CO-EVO rectifies the severe camera bias, reducing the mean cosine distance for same-identity pairs to 0.35, which represents a 48.5\% reduction compared with SSCU. Simultaneously, it expands the margin for different-identity pairs to 0.78, an 85.7\% improvement. This shift ensures that intrinsic identity cues override stylistic shortcuts.
The representation capability in the feature space is further elucidated via the t-SNE visualization in Figure~\ref{fig:visual_evidence}(b), which displays the sample distribution on the unseen MS target domain under the M+C2+C3 to MS setting. In the DACS and SSCU baseline models, identity clusters are scattered and heavily interleaved, indicating that the encoder fails to distinguish between different individuals under significant domain shifts. In contrast, CO-EVO successfully reconciles these scattered samples into compact and well-separated clusters as indicated by the arrows. This qualitative evidence confirms that our CSA anchors act as stable gravitational centers, pulling stylistically diverse inputs toward domain-agnostic semantic centers and thereby establishing a robust decision boundary on novel domains.
Qualitatively, Figure~\ref{fig:visual_evidence}(c) compares our GSD with the standard Style Transformation Model (STM) used in baselines. Notably, GSD maintains superior stylization diversity and image quality across both early and late training stages. While STM-generated images frequently suffer from repetitive artifacts or unrealistic over-exposure in the late stage, our GCSB-based GSD module synthesizes diverse yet realistic camera effects such as varying illumination and color tones. This stability throughout the entire training process ensures that the model consistently learns from high-fidelity proxies of unseen imaging pipelines, effectively enhancing the generalization to novel domains.

\section{Conclusion}
\label{sec:conclusion}
In this paper, we identify and resolve the semantic-style conflict in FedDG-ReID. We propose CO-EVO, a novel framework that bridges the gap between semantic stability and visual diversity through a coupled semantic--style optimization mechanism. By introducing CSA, we establish purified, domain-agnostic identity prototypes that effectively filter out local camera noise via cross-camera consistency. Simultaneously, our GSD module utilizes a privacy-preserving camera-style bank to synthesize realistic domain shifts, expanding the visual boundaries of training data with negligible overhead. Extensive experiments, including analyses with noisy and missing metadata, demonstrate that CO-EVO achieves SOTA performance across multiple benchmarks. These results show that stable semantic grounding and realistic style expansion are jointly essential for robust cross-domain generalization in federated environments.
\section{Limitations}
Despite the performance gains achieved by CO-EVO, certain limitations remain to be addressed in future research. 
First, while CSA provides robust semantic grounding by distilling identity-specific features, these anchors are primarily derived from source-domain identities. Under extreme conditions such as severe occlusion, significantly low resolution, or out-of-distribution appearances not covered by the source distribution, the efficacy of semantic anchoring as a regularizer may diminish. 
Second, the GCSB represents domain shifts using channel-wise statistics. Although this mechanism is lightweight and controllable compared to learning-based generators, it may not fully capture complex geometric transformations, background structural variations, or intricate occlusion patterns inherent in certain imaging pipelines. 
Third, the construction of style templates relies on the availability of camera identifiers or reliable grouping metadata to summarize statistics. In scenarios where such metadata is missing or highly noisy, the fidelity and diversity of the synthesized style bank can degrade, potentially leading to suboptimal generalization. 
Finally, although CO-EVO avoids the heavy computational burden of training generative networks and adheres to the decentralized nature of federated learning, it still introduces extra local computation for generating stylized views and optimizing textual prototypes. Furthermore, while sharing low-dimensional statistics minimizes privacy leakage, we do not provide a formal privacy guarantee such as differential privacy or secure aggregation for the shared camera footprint. 
Exploring more comprehensive style representations and integrating formal privacy-preserving mechanisms remain promising directions for future work.

\bibliography{custom}

\newpage
\appendix

\section{Appendix}

\subsection{A. Implementation Details and Evaluation Protocols}

\paragraph{Evaluation Protocols.}
To comprehensively evaluate both source-domain performance and generalization to unseen target domains, we adopt three evaluation protocols. 
(i) Protocol I (Leave-one-out): We follow a leave-one-domain-out scheme over {C2, C3, MS, M}. In each run, one dataset is held out for testing and the remaining three datasets serve as source domains for federated training. 
(ii) Protocol II (Robustness): As a complement to Protocol I, we further examine the robustness of our method under a smaller number of participating clients by removing one additional source dataset when MS is included in training. 
(iii) Protocol III (Source-domain evaluation): We evaluate the learned model on source domains by testing on their corresponding test splits, which reflects performance in the federated source-domain setting. For all protocols, performance is measured by Mean Average Precision (mAP) and Cumulative Matching Characteristic (CMC) at Rank-1 (R1).

\paragraph{Federated Setup and Hyperparameters.}
We treat each source domain as an individual client and train the shared ReID model under the standard federated learning paradigm. The training process consists of two stages. First, we conduct a semantic learning phase for 120 rounds to initialize stable identity anchors. Subsequently, we run 60 communication rounds for the federated optimization phase. In each round, the server broadcasts the global parameters and each client performs local optimization for E=1 epoch before uploading model updates for aggregation. Based on our hyperparameter analysis, the token length L in CSA is set to 4 to provide the optimal balance between discriminative power and anchoring stability. The cross-camera consistency weight $\lambda_{c3}$ is set to 0.1 to effectively purify semantic anchors by filtering out local camera specific noise. The temperature parameter $\tau$ used in the semantic alignment loss is set to 0.07.

\paragraph{Implementation Details.}
During federated training, the mini-batch size is set to 64. We use the SGD optimizer with an initial learning rate of 1e-3, weight decay of 5e-4, and momentum of 0.9. The learning rate is scheduled by MultiStepLR with decay factors at specific communication rounds. Our framework is compatible with various backbones, including ResNet-50 (RN50) and Vision Transformer (ViT). For the vision-language component, we employ the CLIP model with a ViT-B/16 image encoder as the base for semantic anchoring. For GSD, the camera-style templates are extracted once at the beginning of Phase II to construct the GCSB, which takes about 4s per client on average and contributes less than $0.1\%$ of the full training time. This ensures that the diversification process is grounded in real camera distributions without requiring expensive generators or incurring additional communication overhead in subsequent training rounds.

\end{document}